\documentclass[11pt,a4paper]{article}
\usepackage{authblk}
\usepackage[hyperref]{acl2019}
\usepackage{times}
\usepackage{latexsym}
\usepackage{xcolor}
\usepackage{color}
\usepackage{url}

\aclfinalcopy 
\def\aclpaperid{1510} 

\usepackage{verbatim}
\usepackage{graphicx}
\usepackage{tabularx}
\usepackage{booktabs}
\usepackage{url}
\usepackage{dsfont}
\usepackage{todonotes}
\usepackage{amsmath,amsfonts,amssymb}
\usepackage{subfigure}

\definecolor{mypink1}{rgb}{0.858,0.188,0.478}
\definecolor{mygreen}{rgb}{0.0,0.5,0.0}

\definecolor{carnelian}{rgb}{0.8, 0.0, 0.0}
\definecolor{ceruleanblue}{rgb}{0.0, 0.45, 0.73}

\title{Making Fast Graph-based Algorithms with Graph Metric Embeddings}

\author[$\dag$]{\textbf{Andrey Kutuzov}}
\author[$\ddag$]{\textbf{Mohammad Dorgham}}
\author[$\ddag$]{\textbf{Oleksiy Oliynyk}}
\author[$\ddag$]{\\\textbf{Chris Biemann}}
\author[$\star,\ddag$]{\textbf{Alexander Panchenko}}
\affil[$\dag$]{Language Technology Group, University of Oslo, Oslo, Norway}
\affil[$\ddag$]{Language Technology Group, Universit{\"at} Hamburg, Hamburg, Germany}
\affil[$\star$]{Skolkovo Institute of Science and Technology, Moscow, Russia}

\begin{document}
\maketitle
\begin{abstract}

The computation of distance measures between nodes in graphs is inefficient and does not scale to large graphs. We explore dense vector representations as an effective way to approximate the same information: we introduce a simple yet efficient and effective approach for learning graph embeddings. Instead of directly operating on the graph structure, our method takes structural measures of pairwise node similarities into account and learns dense node representations reflecting  user-defined graph distance measures, such as e.g.~the shortest path distance or distance measures that take information beyond the graph structure into account. We demonstrate a speed-up of several orders of magnitude when predicting word similarity by vector operations on our embeddings as opposed to directly computing the respective path-based measures, while outperforming various other graph embeddings on semantic similarity and word sense disambiguation tasks and show evaluations on the WordNet graph and two knowledge base graphs.
\end{abstract}

When operating on large graphs, such as transportation networks, social networks, or lexical resources, the need for estimating similarities between nodes arises. For many domain-specific applications, custom graph node similarity measures $\mathop{sim}: V\times V \rightarrow \mathds{R}$ have been defined on pairs of nodes $V$ of a graph $G = (V, E)$. Examples include travel time, communities, or semantic distances for knowledge-based word sense disambiguation on WordNet \cite{wordnet:1993}. For instance, the similarity $s_{ij}$ between the {\small \textsf{cup.n.01}} and {\small \textsf{mug.n.01}} synsets in the WordNet is $\frac{1}{4}$ according to the inverted shortest path distance  as these two nodes are connected by the undirected path {\small \textsf{cup} $\rightarrow$ \textsf{container} $\leftarrow$ \textsf{vessel} $\leftarrow$ \textsf{drinking\_vessel} $\leftarrow$ \textsf{mug}}. 

In recent years, a large variety of such node similarity measures have been described, many of which are based on the notion of a random walk ~\cite{fouss2007random,pilehvar2015senses,lebichot2018constrained}. As given by the structure of the problem, most such measures are defined as traversals of edges $E$ of the graph, which makes their computation prohibitively inefficient.

To this end, we propose the \textit{path2vec} model\footnote{\url{https://github.com/uhh-lt/path2vec}}, which solves this problem by decoupling development and use of graph-based measures, and -- in contrast to purely walk-based embeddings -- is trainable to reflect custom node similarity measures. We represent nodes in a graph with dense embeddings that are good in approximating such custom, e.g. application-specific, pairwise node similarity measures. Similarity computations in a vector space are several orders of magnitude faster than computations directly operating on the graph.

First, effectiveness of our model is shown \textit{intrinsically}  by learning metric embeddings for three types of graphs (WordNet, FreeBase, and DBPedia),
based on several similarity measures. Second, in an \textit{extrinsic} evaluation on the Word Sense Disambiguation (WSD) task~\cite{navigli2009word} we replace several original measures with their vectorized counterparts in a known graph-based WSD algorithm by \citet{sinha2007unsupervised}, reaching comparable levels of performance with the graph-based algorithms while maintaining computational gains. 

The main contribution of this paper is the demonstration of the effectiveness and efficiency of the \textit{path2vec} node embedding method~\cite{kutuzov-etal-2019-learning}. This method learns dense vector embeddings of nodes $V$ based on a user-defined custom similarity measure $sim$, e.g. the shortest path distance or any other similarity measure. While our method is able to closely approximate quite different similarity measures as we show on WordNet-based measures and therefore can be used in lieu of these measures in NLP components and applications, our main point is the increase of speed in the similarity computation of nodes, which gains up to 4 orders of magnitude with respect to the original graph-based algorithms.

\section{Graph Metric Embeddings Model}
\label{subsec:model}

\paragraph{Definition of the Model} 

\textit{Path2vec} learns embeddings of the graph nodes $\{v_i,v_j\} \in V$ such that the dot products between pairs of the respective vectors $(\mathbf{v}_i \cdot \mathbf{v}_j)$ are close to the user-defined similarities between the nodes $s_{ij}$. In addition, the model reinforces the similarities $\mathbf{v}_i \cdot \mathbf{v}_{n}$ and $ \mathbf{v}_j \cdot \mathbf{v}_{m}$ between the nodes $v_i$ and $v_j$ and all their respective adjacent nodes $\{v_{n}: \exists (v_i, v_{n})\in E\}$ and $\{v_{m}: \exists (v_j, v_{m})\in E\}$ to preserve local structure of the graph. The model preserves both \textcolor{carnelian}{{global}} and \textcolor{ceruleanblue}{{local}} relations between nodes by minimizing 
$\sum_{ (v_i, v_j) \in B }((\textcolor{carnelian}{\mathbf{v}_i^\top \mathbf{v}_j - s_{ij}} )^2 - \alpha (\textcolor{ceruleanblue}{\mathbf{v}_i^\top \mathbf{v}_{n}} +  \textcolor{ceruleanblue}{\mathbf{v}_j^\top \mathbf{v}_{m}})),
$ 
where $s_{ij} = \mathop{sim}(v_i, v_j)$ is the value of a `gold' similarity measure between a pair of nodes $v_i$ and $v_j$,  $\mathbf{v}_i$ and  $\mathbf{v}_j$ are the embeddings of the first and the second node, $B$ is a training batch, 
$\alpha$ is a regularization coefficient. The second term ($\textcolor{ceruleanblue}{\mathbf{v}_i \cdot \mathbf{v}_{n}} +  \textcolor{ceruleanblue}{\mathbf{v}_j \cdot \mathbf{v}_{m}}$) in the objective function is a regularizer that aids the model to simultaneously maximize the similarity between adjacent nodes while learning the similarity between the two target nodes (one adjacent node is randomly sampled for each target node).

We use negative sampling to form a training batch $B$  adding $p$ negative samples ($s_{ij}=0$) for each  real ($s_{ij} > 0$) training instance: each real node (synset) pair $(v_i, v_j)$ with `gold' similarity $s_{ij}$ is accompanied with $p$ `negative' node pairs $(v_i, v_k)$ and $(v_j, v_l)$ with zero similarities, where $v_k$ and $v_l$ are randomly sampled nodes from $V$. Embeddings are initialized randomly and trained using the  \textit{Adam} optimizer \cite{kingma2014adam} with early stopping. %
Once the model is trained, the computation of node similarities is approximated with the dot product of the learned node vectors, making the computations efficient: $\hat{s}_{ij} = \mathbf{v}_i \cdot \mathbf{v}_j$. 

\paragraph{Relation to Similar Models} Our model bears resemblance to the Skip-gram model \cite{Mikolov:2013}, where the vector dot product $\mathbf{v}_i \cdot  \mathbf{\Tilde{v}}_j$ of vectors of pairs of words $(v_i, v_j)$ from a training corpus is optimized to a high score close to 1 for observed samples, while the dot products of negative samples are optimized towards~0.   In the Skip-gram model, the target is to minimize the log likelihood of the conditional probabilities of context words $w_j$ given current words $w_i$: $\mathcal{L} = - \sum_{ (v_i, v_j) \in B_p } \log \sigma{( \textcolor{ceruleanblue}{\mathbf{v}_i \cdot \mathbf{\Tilde{v}}_j}  )  } - \sum_{ (v_i, v_j) \in B_n } \log \sigma{(  \textcolor{ceruleanblue}{ \mathbf{-v}_i \cdot \mathbf{\Tilde{v}}_j}   )}$,
where $B_p$ is the batch of positive training samples, $B_n$ is the batch of the generated negative samples, and $\sigma$ is the sigmoid function. At this, Skip-gram uses only \textcolor{ceruleanblue}{local} information, never creating the full co-occurrence count matrix. In our \textit{path2vec} model, the target dot product values $s_{ij}$ are not binary, but can take arbitrary values in the $[0...1]$ range, as given by the custom distance metric. Further, we use only a single embedding matrix with vector representations of the graph nodes, not needing to distinguish target and context.

Another related model is Global Vectors (GloVe) \cite{pennington2014glove}, which learns co-occurrence probabilities in a given corpus. The objective function to be minimized in GloVe model is $\mathcal{L} = \sum_{ (v_i, v_j) \in B } f( \textcolor{carnelian}{ s_{ij} } ) (  \textcolor{carnelian}{ {\mathbf{v}_i \cdot \mathbf{\Tilde{v}}_j - \log s_{ij} } } + b_i + b_j  )^2$, where $s_{ij}$ counts the co-occurrences of words $v_i$ and $v_j$, $b_i$ and $b_j$ are additional biases for each word, and $f(s_{ij})$ is a weighting function handling rare co-occurrences. Like the Skip-gram, GloVe also uses two embedding matrices, but it relies only on \textcolor{carnelian}{global} information, pre-aggregating global word co-occurrence counts.

\paragraph{Computing Training Similarities}
\label{sec:datasets}

In general case, our model requires computing pairwise node similarities $s_{ij}$ for training between all pairs of nodes in the input graph $G$. This step could be computationally expensive, but it is done only once to make computing of similarities fast. Besides, for some metrics, effective algorithms exist that compute all pairwise similarities at once, e.g. \citet{johnson1977efficient} algorithm for computing shortest paths distances with the worst-case performance of $O(|V|^2 \log|V| + |V||E|)$. As the input training dataset also grows quadratically in $|V|$, training time for large graphs can be slow. To address this issue, we found it useful to prune the input training set so that each node $v_i \in V$ has only $k \in [50;200]$ most similar nodes. Such pruning does not lead to loss of effectiveness.

\begin{figure}[ht]
    \centering
    \includegraphics[scale=0.66,keepaspectratio]{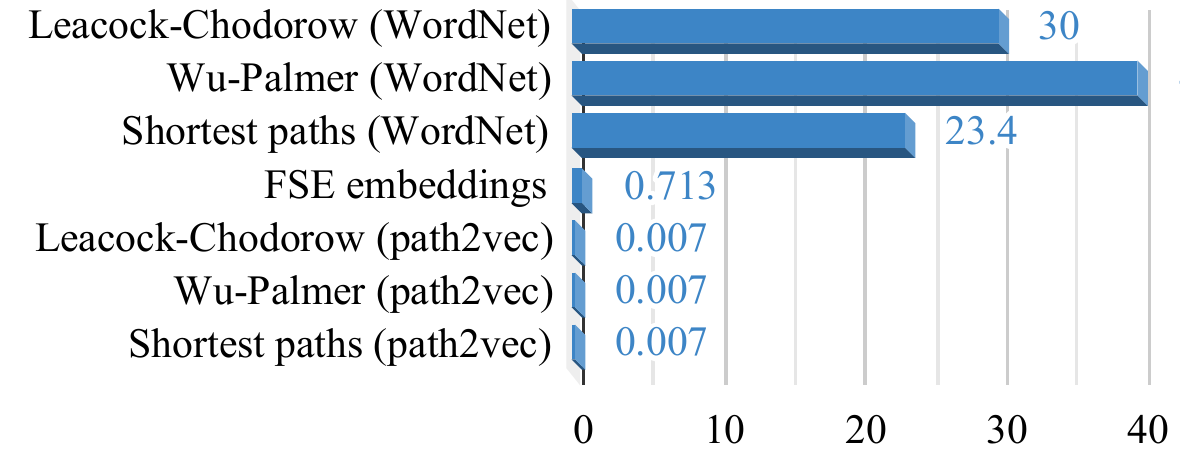}
    \includegraphics[scale=0.67,keepaspectratio]{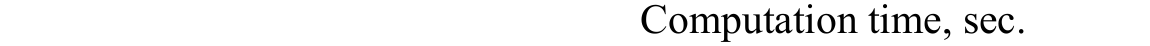}

       \caption{Similarity computation: graph vs vectors.}
       \label{fig:speedup}
\vspace{-15pt}
\end{figure}

\section{Computational Efficiency}

\paragraph{Experimental Setting}

In this section, we compare efficiency of our method as compared to  the original graph based similarity metrics. We trained the model on a graph of 82,115 noun synsets from WordNet. Using NLTK \cite{bird2009natural} we computed the following metrics: (1) Leacock-Chodorow similarities (\textit{LCH}) based on the shortest path between two synsets in the WordNet hypernym/hyponym taxonomy and its maximum depth; (2) inverted shortest path distance (\textit{ShP}); (3) Wu-Palmer similarities (\textit{WuP})  based on the depth of the two nodes in the taxonomy and the depth of their most specific ancestor node. For instance, for \textit{LCH} this procedure took about 30 hours on an Intel Xeon E5-2603v4@1.70GHz CPU using 10 threads. We pruned similarities to the first 50 most similar `neighbors' of each synset and trained \textit{path2vec} on this dataset.

\paragraph{Discussion of Results}

Figure \ref{fig:speedup} presents computation times for pairwise similarities between one synset and all other 82,115 WordNet noun synsets. We compare running times of calculating two original graph-based metrics to Hamming distance between 128D FSE binary embeddings~\cite{subercaze:2015} and to dot product between their dense vectorized 300D counterparts (using CPU).  Using float vectors (\textit{path2vec}) is 4 orders of magnitude faster than operating directly on graphs, and 2 orders faster than Hamming distance. The dot product computation is much faster as compared to shortest path computation (and other complex walks) on a large graph. Also, low-dimensional vector representations of nodes take much less space than the pairwise similarities between all the nodes. The time complexity of calculating the shortest path between graph nodes (as in \textit{ShP} or \textit{LCH}) is in the best case linear in the number of nodes and edges. Calculating Hamming distance between binary strings is linear in the sum of string lengths, which are equivalent of vector sizes \cite{hamming1950error}. At the same time, the complexity of calculating dot product between float vectors is linear in the vector size and is easily parallelized.

\section{Evaluation on Semantic Similarity}
\label{sec:experimental}

\paragraph{Experimental Setting}
We use noun pairs from the SimLex999 dataset \cite{Hill2015}, measuring Spearman rank correlation between `gold' WordNet distances for these pairs and the vector distances produced by the graph embedding models (trained on WordNet) to see how well the models fit the training objective.
We also test the plausibility of the model's output to human judgments. For this, we use human-annotated similarities from the same SimLex999. Some SimLex999 lemmas can be mapped to more than one WordNet synset. We chose the synset pair with the  highest dot product between the embeddings from the corresponding model. 

\paragraph{Baselines}
Our model is compared against five baselines: {\em raw WordNet similarities} by respective measures; {\em DeepWalk} \cite{perozzi2014deepwalk}; {\em node2vec} \cite{grover2016node2vec}; \textit{FSE} \cite{subercaze:2015}; {\em TransR} \cite{transr:2015}. \textit{DeepWalk}, \textit{node2vec}, and \textit{TransR} models were trained on the same WordNet graph. We used all 82,115 noun synsets as vertices and hypernym/hyponym relations between them as edges. 
During the training of \textit{DeepWalk} and \textit{node2vec} models, we tested different values for the number of random walks (in the range from 10 to 100), and the vector size (100 to 600). For \textit{DeepWalk}, we additionally experimented with the window size (5 to 100). All other hyperparameters were left at default values. \textit{FSE} embeddings of  the WordNet noun synsets were provided to us by the authors, and consist of 128-bit vectors.

\begin{table}

\footnotesize
\centering 
\begin{tabular}{l|ccc|ccc}
\toprule
 & LCH & ShP & WuP & LCH & ShP & WuP \\
\midrule
WordNet & \textit{100} & \textit{100} & \textit{100} & 51.3 & 51.3 & 47.4 \\
path2vec  & \textbf{93.5} & \textbf{95.2} & \textbf{93.1} & \textbf{53.2} & \textbf{55.5} & \textbf{55.5} \\

\midrule
TransR & 77.6 & 77.6 & 72.5 & \multicolumn{3}{c}{ 38.6 } \\
node2vec & 75.9 & 75.9 & 78.7  & \multicolumn{3}{c}{ 46.2 } \\ 
DeepWalk & 86.8 & 86.8 & 85.0 & \multicolumn{3}{c}{ 53.3 } \\
FSE & 90.0 & 90.0 & 89.0 & \multicolumn{3}{c}{ 55.6 } \\
\bottomrule
\end{tabular}
\caption {Spearman correlations with WordNet similarities (left) and  human judgments (right) $\times 100$.} 
\label{tab:eval_wordnet}
\end{table}

\begin{figure*}[ht]
    \centering
      \includegraphics[scale=0.355,keepaspectratio]{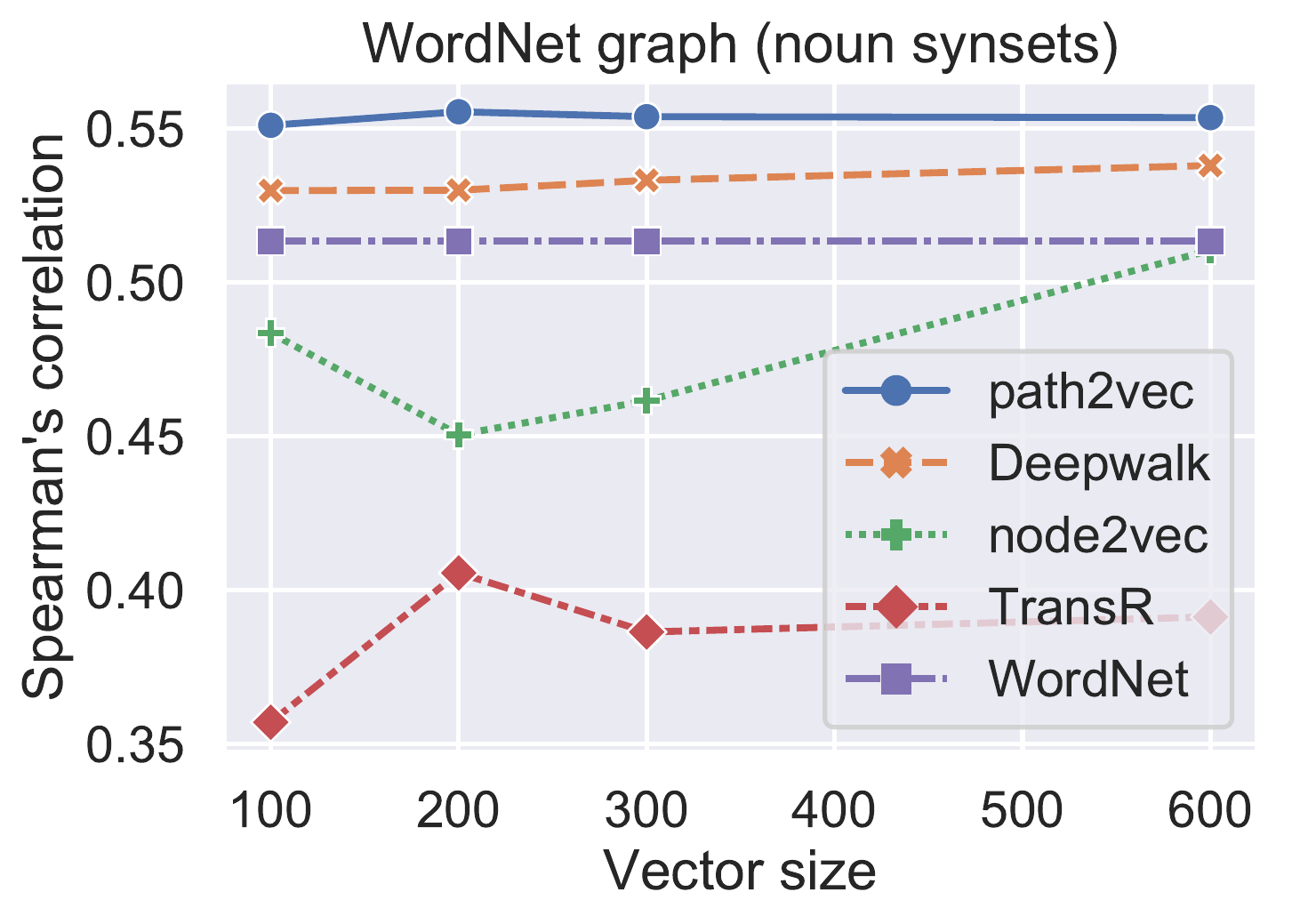}
      \includegraphics[scale=0.355,keepaspectratio]{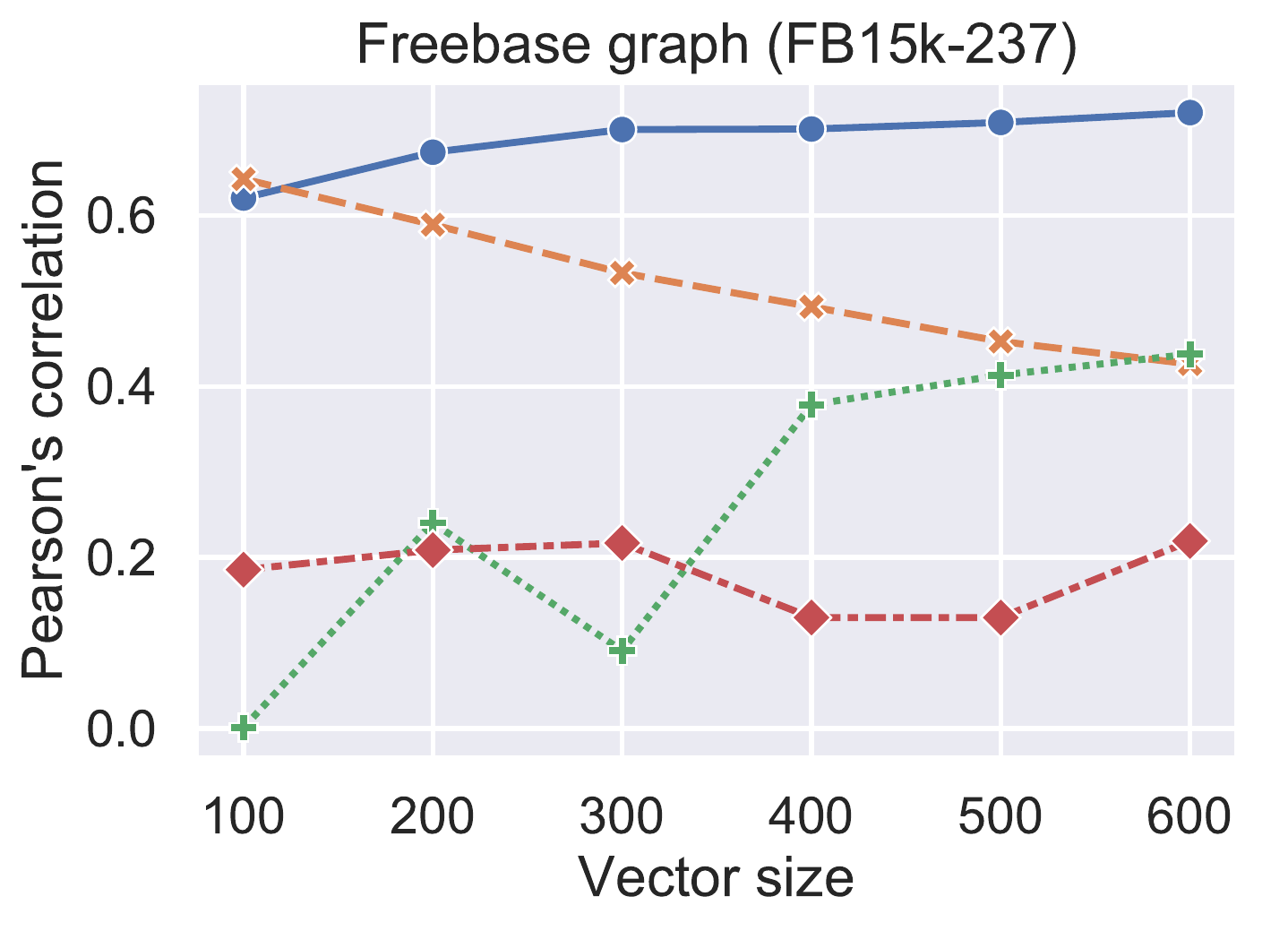}
      \includegraphics[scale=0.355,keepaspectratio]{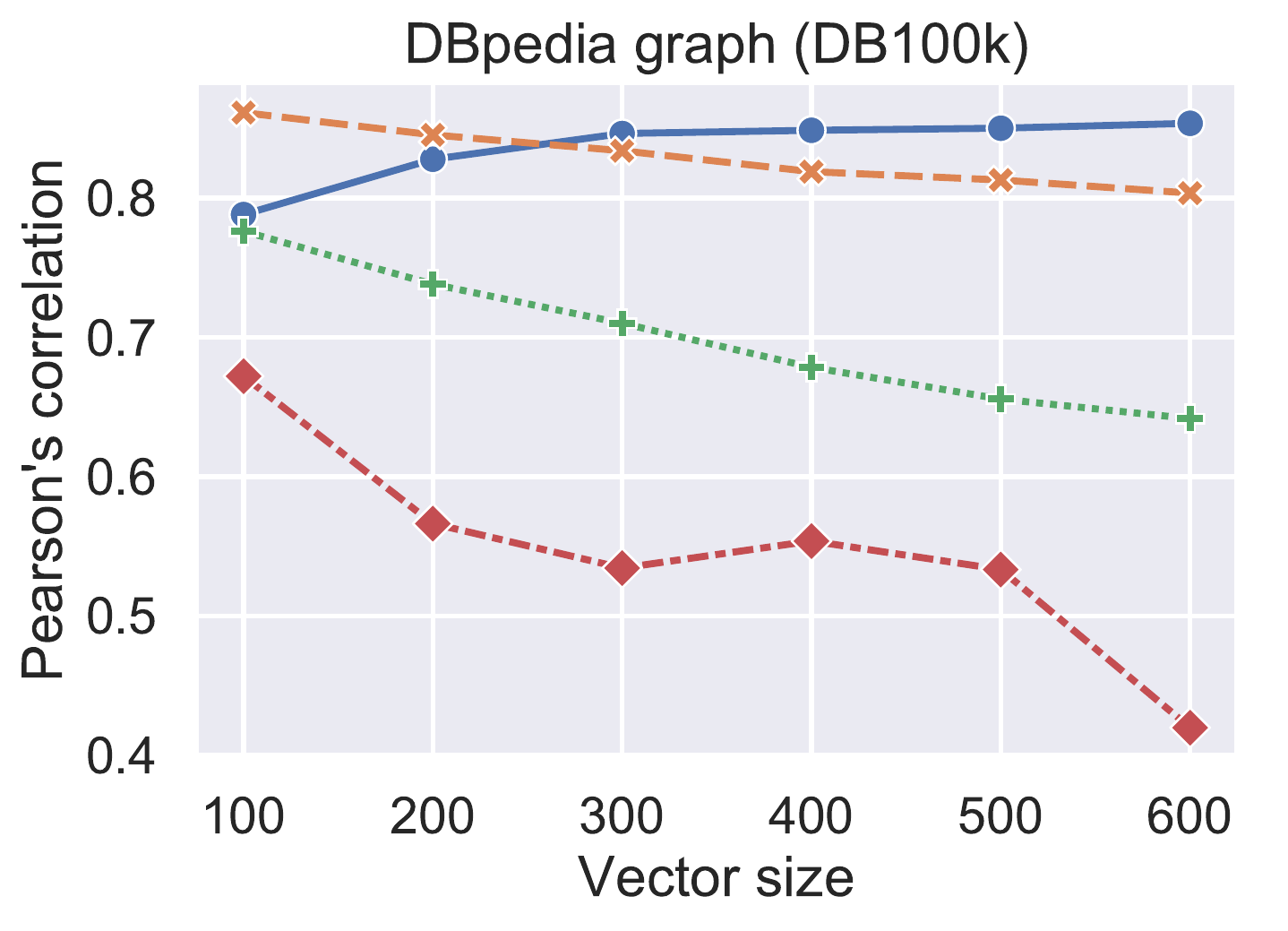}

      \caption{Evaluation on different graphs on SimLex999 (left) and shortest path distance (middle, right).}
      \label{fig:plots}
\end{figure*}

\paragraph{Discussion of Results}
\label{sec:results}

The left part of Table~\ref{tab:eval_wordnet} shows results with the WordNet similarity scores used as gold standard.  \textit{Path2vec} outperforms other graph embeddings, achieving high correlations with WordNet similarities. This shows that our model efficiently approximates  different graph measures. 
The right part of Table~\ref{tab:eval_wordnet} shows results for the correlations with human judgments (SimLex999). We report the results for the best models for each method, all of them (except \textit{FSE}) using vector size 300 for comparability. 

Figure \ref{fig:plots} (left) compares \textit{path2vec} to the baselines, as measured by the correlations with SimLex999 human judgments. The WordNet line denotes the correlation of WordNet similarities with SimLex999 scores. For the \textit{path2vec} models, there is a tendency to improve the performance when the vector size is increased (horizontal axis), until a plateau is reached beyond 600. Note that 
\textit{node2vec} fluctuates, yielding low scores for 200 dimensions. The reported best \textit{DeepWalk} models were trained with the 10 walks and window size 70. The reported best \textit{node2vec} models were trained with 25 walks. Interestingly, \textit{path2vec} and \textit{DeepWalk} models consistently \textit{outperform} the raw WordNet.

\section{Evaluation inside a WSD Algorithm}
\label{subsec:wsd}

\paragraph{Experimental Setting}
To showcase how our approach can be be used inside a graph-based algorithm, we employ word sense disambiguation (WSD) task, reproducing the approach  of \cite{sinha2007unsupervised}. We replace graph similarities with the dot product between node embeddings and study how it influences the WSD performance. The WSD algorithm starts with building a graph where the nodes are the WordNet synsets of the words in the input sentence. The nodes are then connected by edges weighted with the similarity values between the synset pairs. The final step is selecting the most likely sense for each word based on the weighted in-degree centrality score for each synset. 

\paragraph{Discussion of Results}
Table \ref{tab:eval_wsd} presents the WSD micro-F1 scores  using raw WordNet similarities, 300D \textit{path2vec}, \textit{DeepWalk} and \textit{node2vec} models, and the 128D \textit{FSE} model. We evaluate on the following all-words English WSD test sets: Senseval-2 \cite{senseval:2001}, Senseval-3 \cite{senseval3:2004}, and SemEval-15 Task 13 \cite{semeval:2015}. The raw WordNet similarities have a small edge over their vector approximations in the majority of the cases yet the \textit{path2vec} models consistently closely follow them while outperforming other graph embedding baselines: We indicate the differences with respect to the original with a subscript number.

\begin{table}

\centering

\scalebox{0.82}{\parbox{\textwidth}{

\begin{tabular}{lccc}
\toprule
\bf Model & \bf  Senseval2 &\bf  Senseval3 & \bf SemEval-15 \\
\midrule

Random sense & 0.381  & 0.312 & 0.393 \\
\midrule

\multicolumn{4}{c}{ \textit{Graph-based vs vector-based measures} } \\
\midrule
LCH (WordNet) & 0.547\textcolor{white}{$_{\downarrow0.000}$} & 0.494\textcolor{white}{$_{\downarrow0.000}$} & 0.550\textcolor{white}{$_{\downarrow0.000}$}  \\
LCH (path2vec) & 0.527\textcolor{red}{$_{\downarrow0.020}$} & 0.472\textcolor{red}{$_{\downarrow0.022}$} & 0.536\textcolor{red}{$_{\downarrow0.014}$}  \\

\midrule 

ShP (WordNet) & 0.548\textcolor{white}{$_{\downarrow0.000}$} & 0.495\textcolor{white}{$_{\downarrow0.000}$} & 0.550\textcolor{white}{$_{\downarrow0.000}$}  \\
ShP (path2vec) & 0.534\textcolor{red}{$_{\downarrow0.014}$} & 0.489\textcolor{red}{$_{\downarrow0.006}$} & 0.563\textcolor{mygreen}{$_{\uparrow0.013}$} \\

\midrule 
WuP (WordNet) & 0.547\textcolor{white}{$_{\downarrow0.000}$} & 0.487\textcolor{white}{$_{\downarrow0.000}$} &  0.542\textcolor{white}{$_{\downarrow0.000}$} \\
WuP (path2vec) & 0.543\textcolor{red}{$_{\downarrow0.004}$} & 0.489\textcolor{mygreen}{$_{\uparrow0.002}$} & 0.545\textcolor{mygreen}{$_{\uparrow0.003}$} \\

\midrule

\multicolumn{4}{c}{ \textit{Various baseline graph embeddings trained on WordNet}} \\
\midrule
TransR  & 0.540  & 0.466 & 0.536  \\
node2vec  & 0.503 & 0.467 & 0.489  \\
DeepWalk  & 0.528 & 0.476 & 0.552  \\
FSE  & 0.536 & 0.476 & 0.523 \\

\bottomrule
\end{tabular}

}}

\caption {F1 scores of a graph-based WSD algorithm on WordNet versus its vectorized counterparts.}
\label{tab:eval_wsd}
\end{table}

\section{Evaluation on Knowledge Base Graphs}

\paragraph{Experimental Settings}

To show the utility of our model besides the WordNet graph, we also applied it to two graphs derived from knowledge bases (KBs). More specifically, we base our experiments on two publicly available standard samples from these two resources: the FB15k-237~\cite{toutanova2015observed} dataset contains 14,951 entities/nodes and is derived from  Freebase~\cite{bollacker2008freebase}; the DB100k~\cite{ding-etal-2018-improving} dataset contains 99,604 entities/nodes and is derived from DBPedia~\cite{auer2007dbpedia}.   

It is important to note that both datasets were used to evaluate approaches that learn knowledge graph embeddings, e.g.  \cite{transr:2015,xie2016representation,joulin2017fast} on the task on \textit{knowledge base completion} (KBC), to predict missing KB edges/relations between nodes/entities. The specificity of our model is that it learns a given graph similarity metric, which is not provided in these datasets. Therefore, we use only the graphs from these datasets, computing the shortest path distances between all pairs of nodes using the algorithm of \citet{johnson1977efficient}. Instead of the KBC task, we evaluate on the task of predicting node similarity, here using the shortest path distance. We generate a random sample of node pairs for testing from the set of all node pairs (these pairs are excluded from training). The test set contains an equal number of paths of length 1-7 (in total 1050 pairs each, 150 pairs per path length).

\paragraph{Discussion of Results}
Figure~\ref{fig:plots} (middle and right) shows evaluation results on the knowledge base graphs. \textit{Path2vec} is able to better approximate the target graph metric than the standard graph embedding models. As dimensionality of the embeddings increases, the model more closely approximates the target metric, but the performance drop for the models with a low number of dimensions is not drastic, allowing more effective computations while maintaining a reasonable efficiency level. 
Regarding the competitors, DeepWalk comes closest to the performance of our approach, but does not seem to make use of the additional dimensions when training on larger vector sizes; on the DBPedia dataset, this issue is shared between all baselines, where correlation to the true path lengths decreases as representation length increases.

\section{Related Work} \label{sec:related}

Representation learning on graphs received much attention recently in various research communities, see \citet{hamilton2017representation} for a thorough survey on the existing methods. All of them (including ours) are based on the idea of projecting graph nodes into a latent space with a much lower dimensionality than the number of nodes.

Existing approaches to graph embeddings use either factorization of the graph adjacency matrix  \cite{cao2015grarep,ou2016asymmetric} or random walks over the graph as in \textit{Deepwalk} \cite{perozzi2014deepwalk} and \textit{node2vec} \cite{grover2016node2vec}. A different approach is taken by \citet{subercaze:2015}, who directly embed the WordNet tree graph into Hamming hypercube binary representations. Their `Fast similarity embedding' (\textit{FSE}) model provides a quick way of calculating semantic similarities based on WordNet. The \textit{FSE} embeddings are not differentiable though, considerably limiting their use in deep neural architectures. 
\textit{TransR} \cite{transr:2015} extends \textit{TransH} \cite{transh:2014} and is based on the idea that an entity may have a few aspects and different relations are focused on them. So the same entities can be close or far from each other depending on the type of the relation. 
\textit{TransR} projects entity vectors into a relation specific space, and learns embeddings via translation between projected entities. 

We compare our \textit{path2vec} model to these approaches, yet we did not compare to the models like \textit{GraphSAGE} embeddings \cite{hamilton2017inductive} and Graph Convolutional Networks \cite{kipf2018modeling} as they use node features which are absent in our setup.

\section{Conclusion}

Structured knowledge contained in language networks is useful for NLP applications but is difficult to use directly in neural architectures. We proposed a way to train embeddings that directly represent a graph-based similarity measure structure. Our model, \textit{path2vec}, relies on both global and local information from the graph and is simple, effective, and computationally efficient. We demonstrated that our approach generalizes well across graphs (WordNet, Freebase, and DBpedia). Besides, we integrated it into a graph-based WSD algorithm, showing that its vectorized counterpart yields comparable F1 scores on three datasets. 

\textit{Path2vec} enables a speed-up of up to four orders of magnitude for the computation of graph distances as compared to `direct' graph measures. Thus, our model is simple and general, hence it may be applied to any graph
together with a node distance measure to speed up algorithms that employ graph distances.

\subsubsection*{Acknowledgments}
This was supported by the DFG under ``JOIN-T'' (BI 1544/4) and ``ACQuA'' (BI 1544/7) projects. 

\bibliography{acl2019.bib}

\begin{thebibliography}{33}
\expandafter\ifx\csname natexlab\endcsname\relax\def\natexlab#1{#1}\fi

\bibitem[{Auer et~al.(2007)Auer, Bizer, Kobilarov, Lehmann, Cyganiak, and
  Ives}]{auer2007dbpedia}
S{\"o}ren Auer, Christian Bizer, Georgi Kobilarov, Jens Lehmann, Richard
  Cyganiak, and Zachary Ives. 2007.
\newblock \href
  {https://link.springer.com/chapter/10.1007/978-3-540-76298-0_52} {{DB}pedia:
  A nucleus for a web of open data}.
\newblock In \emph{The Semantic Web: Proceedings of the 6th International
  Semantic Web Conference, 2nd Asian Semantic Web Conference, ISWC 2007 + ASWC
  2007}, pages 722--735, Busan, South Korea. Springer.

\bibitem[{Bird et~al.(2009)Bird, Klein, and Loper}]{bird2009natural}
Steven Bird, Ewan Klein, and Edward Loper. 2009.
\newblock \href {https://www.nltk.org/book/} {\emph{Natural language processing
  with Python: analyzing text with the natural language toolkit}}.
\newblock O'Reilly Media, Inc.

\bibitem[{Bollacker et~al.(2008)Bollacker, Evans, Paritosh, Sturge, and
  Taylor}]{bollacker2008freebase}
Kurt Bollacker, Colin Evans, Praveen Paritosh, Tim Sturge, and Jamie Taylor.
  2008.
\newblock \href {https://dl.acm.org/citation.cfm?id=1376746} {Freebase: a
  collaboratively created graph database for structuring human knowledge}.
\newblock In \emph{Proceedings of the 2008 ACM SIGMOD international conference
  on Management of data}, pages 1247--1250, Vancouver, BC, Canada. ACM.

\bibitem[{Cao et~al.(2015)Cao, Lu, and Xu}]{cao2015grarep}
Shaosheng Cao, Wei Lu, and Qiongkai Xu. 2015.
\newblock \href {https://dl.acm.org/citation.cfm?id=2806512} {Gra{R}ep:
  Learning graph representations with global structural information}.
\newblock In \emph{Proceedings of the 24th ACM International on Conference on
  Information and Knowledge Management}, pages 891--900, Melbourne, Australia.
  ACM.

\bibitem[{Ding et~al.(2018)Ding, Wang, Wang, and
  Guo}]{ding-etal-2018-improving}
Boyang Ding, Quan Wang, Bin Wang, and Li~Guo. 2018.
\newblock \href {https://www.aclweb.org/anthology/P18-1011} {Improving
  knowledge graph embedding using simple constraints}.
\newblock In \emph{Proceedings of the 56th Annual Meeting of the Association
  for Computational Linguistics (Volume 1: Long Papers)}, pages 110--121,
  Melbourne, Australia. Association for Computational Linguistics.

\bibitem[{Fouss et~al.(2007)Fouss, Pirotte, Renders, and
  Saerens}]{fouss2007random}
Francois Fouss, Alain Pirotte, Jean-Michel Renders, and Marco Saerens. 2007.
\newblock \href {https://ieeexplore.ieee.org/document/4072747} {Random-walk
  computation of similarities between nodes of a graph with application to
  collaborative recommendation}.
\newblock \emph{IEEE Transactions on knowledge and data engineering},
  19(3):355--369.

\bibitem[{Grover and Leskovec(2016)}]{grover2016node2vec}
Aditya Grover and Jure Leskovec. 2016.
\newblock \href {https://dl.acm.org/citation.cfm?id=2939754} {Node2vec:
  Scalable feature learning for networks}.
\newblock In \emph{Proceedings of the 22nd ACM SIGKDD international conference
  on Knowledge discovery and data mining}, pages 855--864, San Francisco, CA,
  USA. ACM.

\bibitem[{Hamilton et~al.(2017{\natexlab{a}})Hamilton, Ying, and
  Leskovec}]{hamilton2017representation}
William Hamilton, Rex Ying, and Jure Leskovec. 2017{\natexlab{a}}.
\newblock \href {http://sites.computer.org/debull/A17sept/p52.pdf}
  {Representation learning on graphs: Methods and applications}.
\newblock \emph{IEEE Data Engineering Bulletin}, 40(3):52--74.

\bibitem[{Hamilton et~al.(2017{\natexlab{b}})Hamilton, Ying, and
  Leskovec}]{hamilton2017inductive}
William Hamilton, Zhitao Ying, and Jure Leskovec. 2017{\natexlab{b}}.
\newblock \href
  {https://papers.nips.cc/paper/6703-inductive-representation-learning-on-large-graphs.pdf}
  {Inductive representation learning on large graphs}.
\newblock In \emph{Advances in Neural Information Processing Systems}, pages
  1024--1034, Long Beach, CA, USA.

\bibitem[{Hamming(1950)}]{hamming1950error}
Richard Hamming. 1950.
\newblock \href {https://ieeexplore.ieee.org/document/6772729} {Error detecting
  and error correcting codes}.
\newblock \emph{Bell System technical journal}, 29(2):147--160.

\bibitem[{Hill et~al.(2015)Hill, Reichart, and Korhonen}]{Hill2015}
Felix Hill, Roi Reichart, and Anna Korhonen. 2015.
\newblock \href {https://www.aclweb.org/anthology/J15-4004} {{SimLex-999:
  Evaluating Semantic Models With (Genuine) Similarity Estimation}}.
\newblock \emph{Computational {L}inguistics}, 41(4):665--695.

\bibitem[{Johnson(1977)}]{johnson1977efficient}
Donald~B. Johnson. 1977.
\newblock \href {https://dl.acm.org/citation.cfm?id=321993} {Efficient
  algorithms for shortest paths in sparse networks}.
\newblock \emph{Journal of the ACM (JACM)}, 24(1):1--13.

\bibitem[{Joulin et~al.(2017)Joulin, Grave, Bojanowski, Nickel, and
  Mikolov}]{joulin2017fast}
Armand Joulin, Edouard Grave, Piotr Bojanowski, Maximilian Nickel, and Tomas
  Mikolov. 2017.
\newblock \href {https://arxiv.org/abs/1710.10881} {Fast linear model for
  knowledge graph embeddings}.
\newblock \emph{arXiv preprint arXiv:1710.10881}.

\bibitem[{Kingma and Ba(2015)}]{kingma2014adam}
Diederik~P. Kingma and Jimmy Ba. 2015.
\newblock \href {https://hdl.handle.net/11245/1.505367} {Adam: A method for
  stochastic optimization}.
\newblock In \emph{Proceedings of the International Conference on Learning
  Representations (ICLR)}, San Diego, CA, USA.

\bibitem[{Kutuzov et~al.(2019)Kutuzov, Dorgham, Oliynyk, Biemann, and
  Panchenko}]{kutuzov-etal-2019-learning}
Andrey Kutuzov, Mohammad Dorgham, Oleksiy Oliynyk, Chris Biemann, and Alexander
  Panchenko. 2019.
\newblock \href {https://www.aclweb.org/anthology/S19-1014} {Learning graph
  embeddings from {W}ord{N}et-based similarity measures}.
\newblock In \emph{Proceedings of the Eighth Joint Conference on Lexical and
  Computational Semantics (*{SEM} 2019)}, pages 125--135, Minneapolis, MN, USA.
  Association for Computational Linguistics.

\bibitem[{Lebichot et~al.(2018)Lebichot, Guex, Kivim{\"a}ki, and
  Saerens}]{lebichot2018constrained}
Bertrand Lebichot, Guillaume Guex, Ilkka Kivim{\"a}ki, and Marco Saerens. 2018.
\newblock \href {https://arxiv.org/abs/1807.04551} {A constrained randomized
  shortest-paths framework for optimal exploration}.
\newblock \emph{arXiv preprint arXiv:1807.04551}.

\bibitem[{Lin et~al.(2015)Lin, Liu, Sun, Liu, and Zhu}]{transr:2015}
Yankai Lin, Zhiyuan Liu, Maosong Sun, Yang Liu, and Xuan Zhu. 2015.
\newblock \href
  {https://www.aaai.org/ocs/index.php/AAAI/AAAI15/paper/view/9571} {Learning
  entity and relation embeddings for knowledge graph completion}.
\newblock In \emph{Proceedings of the 29th AAAI Conference on Artificial
  Intelligence}, pages 2181--2187, Austin, TX, USA. AAAI Press.

\bibitem[{Mihalcea et~al.(2004)Mihalcea, Chklovski, and
  Kilgarriff}]{senseval3:2004}
Rada Mihalcea, Timothy Chklovski, and Adam Kilgarriff. 2004.
\newblock \href {http://aclweb.org/anthology/W04-0807} {The {S}enseval-3
  {E}nglish lexical sample task}.
\newblock In \emph{Senseval-3: Third International Workshop on the Evaluation
  of Systems for the Semantic Analysis of Text}, pages 25--28, Barcelona,
  Spain. Association for Computational Linguistics.

\bibitem[{Mikolov et~al.(2013)Mikolov, Sutskever, Chen, Corrado, and
  Dean}]{Mikolov:2013}
Tomas Mikolov, Ilya Sutskever, Kai Chen, Greg~S. Corrado, and Jeff Dean. 2013.
\newblock \href
  {https://papers.nips.cc/paper/5021-distributed-representations-of-words-and-phrases-and-their-compositionality.pdf}
  {Distributed representations of words and phrases and their
  compositionality}.
\newblock In \emph{Advances in Neural Information Processing Systems 26}, pages
  3111--3119, Lake Tahoe, NV, USA. Curran Associates, Inc.

\bibitem[{Miller(1995)}]{wordnet:1993}
George~A. Miller. 1995.
\newblock \href {https://doi.org/10.1145/219717.219748} {Word{N}et: A lexical
  database for {E}nglish}.
\newblock \emph{Communications of the ACM}, 38(11):39--41.

\bibitem[{Moro and Navigli(2015)}]{semeval:2015}
Andrea Moro and Roberto Navigli. 2015.
\newblock \href {https://doi.org/10.18653/v1/S15-2049} {Semeval-2015 task 13:
  Multilingual all-words sense disambiguation and entity linking}.
\newblock In \emph{Proceedings of the 9th International Workshop on Semantic
  Evaluation (SemEval 2015)}, pages 288--297. Association for Computational
  Linguistics.

\bibitem[{Navigli(2009)}]{navigli2009word}
Roberto Navigli. 2009.
\newblock \href {https://dl.acm.org/citation.cfm?id=1459355} {Word sense
  disambiguation: A survey}.
\newblock \emph{ACM Computing Surveys (CSUR)}, 41(2):10.

\bibitem[{Ou et~al.(2016)Ou, Cui, Pei, Zhang, and Zhu}]{ou2016asymmetric}
Mingdong Ou, Peng Cui, Jian Pei, Ziwei Zhang, and Wenwu Zhu. 2016.
\newblock \href {https://www.kdd.org/kdd2016/papers/files/rfp0184-ouA.pdf}
  {Asymmetric transitivity preserving graph embedding}.
\newblock In \emph{Proceedings of the 22nd ACM SIGKDD international conference
  on Knowledge discovery and data mining}, pages 1105--1114, San Francisco, CA,
  USA. ACM.

\bibitem[{Palmer et~al.(2001)Palmer, Fellbaum, Cotton, Delfs, and
  Dang}]{senseval:2001}
Martha Palmer, Christiane Fellbaum, Scott Cotton, Lauren Delfs, and Hoa~Trang
  Dang. 2001.
\newblock \href {http://www.aclweb.org/anthology/S01-1005} {English tasks:
  All-words and verb lexical sample}.
\newblock In \emph{Proceedings of SENSEVAL-2 Second International Workshop on
  Evaluating Word Sense Disambiguation Systems}, pages 21--24, Toulouse,
  France. Association for Computational Linguistics.

\bibitem[{Pennington et~al.(2014)Pennington, Socher, and
  Manning}]{pennington2014glove}
Jeffrey Pennington, Richard Socher, and Christopher~D. Manning. 2014.
\newblock \href {http://www.aclweb.org/anthology/D14-1162} {Glove: Global
  vectors for word representation}.
\newblock In \emph{Proceedings of the 2014 Conference on Empirical Methods in
  Natural Language Processing (EMNLP)}, pages 1532--1543, Doha, Qatar.
  Association for Computational Linguistics.

\bibitem[{Perozzi et~al.(2014)Perozzi, Al-Rfou, and
  Skiena}]{perozzi2014deepwalk}
Bryan Perozzi, Rami Al-Rfou, and Steven Skiena. 2014.
\newblock \href {https://dl.acm.org/citation.cfm?doid=2623330.2623732}
  {Deep{W}alk: Online learning of social representations}.
\newblock In \emph{Proceedings of the 20th ACM SIGKDD international conference
  on Knowledge discovery and data mining}, pages 701--710, New York, NY, USA.
  ACM.

\bibitem[{Pilehvar and Navigli(2015)}]{pilehvar2015senses}
Mohammad~T. Pilehvar and Roberto Navigli. 2015.
\newblock \href
  {https://www.sciencedirect.com/science/article/abs/pii/S000437021500106X}
  {From senses to texts: An all-in-one graph-based approach for measuring
  semantic similarity}.
\newblock \emph{Artificial Intelligence}, 228:95--128.

\bibitem[{Schlichtkrull et~al.(2018)Schlichtkrull, Kipf, Bloem, van~den Berg,
  Titov, and Welling}]{kipf2018modeling}
Michael Schlichtkrull, Thomas~N. Kipf, Peter Bloem, Rianne van~den Berg, Ivan
  Titov, and Max Welling. 2018.
\newblock \href
  {https://link.springer.com/chapter/10.1007/978-3-319-93417-4_38} {Modeling
  relational data with graph convolutional networks}.
\newblock In \emph{Proceedings of the European Semantic Web Conference 2018:
  The Semantic Web}, pages 593--607, Heraklion, Greece. Springer.

\bibitem[{Sinha and Mihalcea(2007)}]{sinha2007unsupervised}
Ravi Sinha and Rada Mihalcea. 2007.
\newblock \href {https://ieeexplore.ieee.org/document/4338370} {Unsupervised
  graph-based word sense disambiguation using measures of word semantic
  similarity}.
\newblock In \emph{International Conference on Semantic Computing (ICSC)},
  pages 363--369, Irvine, CA, USA. IEEE.

\bibitem[{Subercaze et~al.(2015)Subercaze, Gravier, and
  Laforest}]{subercaze:2015}
Julien Subercaze, Christophe Gravier, and Fr{\'e}d{\'e}rique Laforest. 2015.
\newblock \href {https://doi.org/10.3115/v1/P15-2002} {On metric embedding for
  boosting semantic similarity computations}.
\newblock In \emph{Proceedings of the 53rd Annual Meeting of the Association
  for Computational Linguistics and the 7th International Joint Conference on
  Natural Language Processing (Volume 2: Short Papers)}, pages 8--14, Beijing,
  China. Association for Computational Linguistics.

\bibitem[{Toutanova and Chen(2015)}]{toutanova2015observed}
Kristina Toutanova and Danqi Chen. 2015.
\newblock \href {https://doi.org/10.18653/v1/W15-4007} {Observed versus latent
  features for knowledge base and text inference}.
\newblock In \emph{Proceedings of the 3rd Workshop on Continuous Vector Space
  Models and their Compositionality}, pages 57--66, Beijing, China. Association
  for Computational Linguistics.

\bibitem[{Wang et~al.(2014)Wang, Zhang, Feng, and Chen}]{transh:2014}
Zhen Wang, Jianwen Zhang, Jianlin Feng, and Zheng Chen. 2014.
\newblock \href
  {https://www.aaai.org/ocs/index.php/AAAI/AAAI14/paper/view/8531} {Knowledge
  graph embedding by translating on hyperplanes}.
\newblock In \emph{Proceedings of the 28th AAAI Conference on Artificial
  Intelligence}, pages 1112--1119, Qu{\'e}bec City, QC, Canada. AAAI Press.

\bibitem[{Xie et~al.(2016)Xie, Liu, Jia, Luan, and Sun}]{xie2016representation}
Ruobing Xie, Zhiyuan Liu, Jia Jia, Huanbo Luan, and Maosong Sun. 2016.
\newblock \href {https://dl.acm.org/citation.cfm?id=3016100.3016273}
  {Representation learning of knowledge graphs with entity descriptions}.
\newblock In \emph{Proceedings of the 30th AAAI Conference on Artificial
  Intelligence}, pages 2659--2665, Phoenix, AZ, USA. AAAI Press.

\end{thebibliography}
\bibliographystyle{acl_natbib}

\end{document}